\documentclass{article}
\usepackage{spconf,amsmath,graphicx,amsfonts}
\usepackage{cite,color,url} 
\newcommand{\myfig}[1]{Fig.\;\ref{#1}}

\newcommand{\mysec}[1]{Section \ref{#1}}

\title{ELBERT: FAST ALBERT WITH CONFIDENCE-WINDOW BASED EARLY EXIT}
\name{Keli Xie, Siyuan Lu, Meiqi Wang, Zhongfeng Wang
\thanks{This work was supported by the National Natural Science Foundation of China under Grant 61774082, the Fundamental Research Funds for the Central Universities under Grant 021014380065, the Key Research Plan of Jiangsu Province of China under Grant BE2019003-4. (Corresponding author: Zhongfeng Wang.)}}
\address{School of Electronic Science and Engineering, Nanjing University, Nanjing, China\\	
			\normalsize{\url{{kelixie, sylu, mqwang}@smail.nju.edu.cn, zfwang@nju.edu.cn}}}

\begin{document}
%
\maketitle
\begin{abstract}
Despite the great success in Natural Language Processing (NLP) area, large pre-trained  language models like BERT are not well-suited for resource-constrained or real-time applications owing to the large number of parameters and slow inference speed. 
Recently, compressing and accelerating BERT have become important topics.
By incorporating a parameter-sharing strategy, ALBERT greatly reduces the number of parameters while achieving competitive performance.
Nevertheless, ALBERT still suffers from a long inference time. In this work, we propose the ELBERT, which significantly improves the average inference speed compared to ALBERT due to the proposed confidence-window based early exit mechanism, without introducing additional parameters or extra training overhead. 
Experimental results show that ELBERT achieves an adaptive inference speedup varying from 2$\times$ to 10$\times$ with negligible accuracy degradation compared to ALBERT on various datasets. 
Besides, ELBERT achieves higher accuracy than existing early exit methods used for accelerating BERT under the same computation cost. 
Furthermore, to understand the principle of the early exit mechanism, we also visualize the decision-making process of it in ELBERT. 
Our code is publicly available online.\footnote{\footnotesize{\url{https://github.com/shakeley/ELBERT}}}
\end{abstract}

\begin{keywords}
  Natural Language Processing, BERT, Inference Acceleration, Early Exit, Model Compression
\end{keywords}
\section{Introduction}
\label{sec:intro} 
In recent years, large pre-trained  language models (e.g., BERT\cite{devlin2019bert}, RoBERTa\cite{liu2019roberta}, XLNet\cite{yang2019xlnet}) have made remarkable improvements in many Natural Language Processing (NLP) tasks. 
However, the great success is achieved at the cost of millions of parameters and huge computation cost. Hence, employing those models in resource-constrained and real-time scenarios is quite difficult. 

To improve the applicability of BERT, some works using common compression methods have been proposed, such as Weight Pruning\cite{han2015deep}, Quantization\cite{gong2014compressing} and Knowledge Distilling\cite{hinton2015distilling}. Compared with the models based on these methods, ALBERT\cite{lan2019albert} greatly reduces the amount of parameters and memory consumption by sharing parameters, and achieves even better performance than BERT. However, ALBERT doesn't cut down computation cost and inference time. 

Redundance\cite{kovaleva2019revealing} and overthinking\cite{kaya2019shallow} are knotty problems that big models often suffer from. 
Early exit is a method that focuses on the differences in input complexities for avoiding redundant computations and overthinking to accelerate inference. 
The inputs judged as simple are processed with only a part of the whole model. Early exit enables one-for-all\cite{cai2019once}, which means that one trained model can meet different accuracy-speed trade-offs by adjusting the criterion of input complexity in inference only, while time-consuming re-training is needed for other common compression methods. 

In this paper, we propose ELBERT, a fast ALBERT coupled with a confidence-window based early exit mechanism, which achieves high-speed inference without introducing additional parameters. 
Specifically, 
ELBERT uses ALBERT as the backbone model (also compatible with other BERT-like models). The confidence-window based early exit mechanism enables an input-adaptive efficient inference. Thus it saves inference time and computation cost. 
We conduct extensive experiments on various datasets. The results show that ELBERT achieves at least 2$\times$ inference speedup while keeping and even improving the accuracy, and up to 10$\times$ speedup with negligible accuracy degradation. 

The main contributions of this paper can be summarized as follows: 1) A novel and efficient confidence-window based early exit mechanism is proposed for the first time to the best of our knowledge. 2) We propose ELBERT which achieves better performance than existing early exit methods used for accelerating BERT on many NLP tasks. 3) We visualize the decision-making process of the early exit in ELBERT, which sheds light on its internal mechanism. 

\section{Related Work}
\label{sec:related}
Prior works in model compression can be mainly divided into two categories:
\\
A. \textbf{\emph{Structure-wise}} compression methods try to remove the unimportant elements of models. For Weight Pruning, Gordon \emph{et al.}\cite{gordon2020compressing} applied the magnitude-based pruning method to BERT. Michel \emph{et al.}\cite{michel2019sixteen} pruned BERT based on gradients of weights. For Quantization, Q-BERT\cite{shen2020q} utilized a Hessian based mix-precision approach to compress BERT, while Q8BERT \cite{zafrir2019q8bert} quantized BERT using symmetric linear quantization. Besides, Knowledge Distilling is applied by Tang \emph{et al.} \cite{tang2019distilling}, Sun \emph{et al.}\cite{sun2019patient}, DistillBERT\cite{sanh2019distilbert} and TinyBERT\cite{jiao2019tinybert} for a light BERT. 
\\
B. \textbf{\emph{Input-wise}} compression methods focus on avoiding redundant computations based on the complexity of inputs. BranchyNet\cite{teerapittayanon2016branchynet} proposed the entropy based confidence measurement. Shallow-Deep Nets\cite{kaya2019shallow} managed to mitigate the overthinking problem with early exit mechanism. LayerDrop\cite{fan2019reducing} randomly dropped layers at training time, allowing for sub-network selection to any desired depth in inference. Concurrently, DeeBERT\cite{xin2020deebert} and TheRT\cite{Schwartz2020TheRT} applied the basic early exit method to BERT. FastBERT\cite{liu2020fastbert} proposed a self-distilling method in fine-tuning. However, those works only explored the intermediate state of the classifier, while ELBERT proposes a two-stage early exit mechanism. Coincidentally, Zhou\cite{Wangchunshu} first proposed one criterion which is similar to one of the proposed criteria in this work. 
  \begin{figure}[t]
    \centering
    \includegraphics[width=8.5cm]{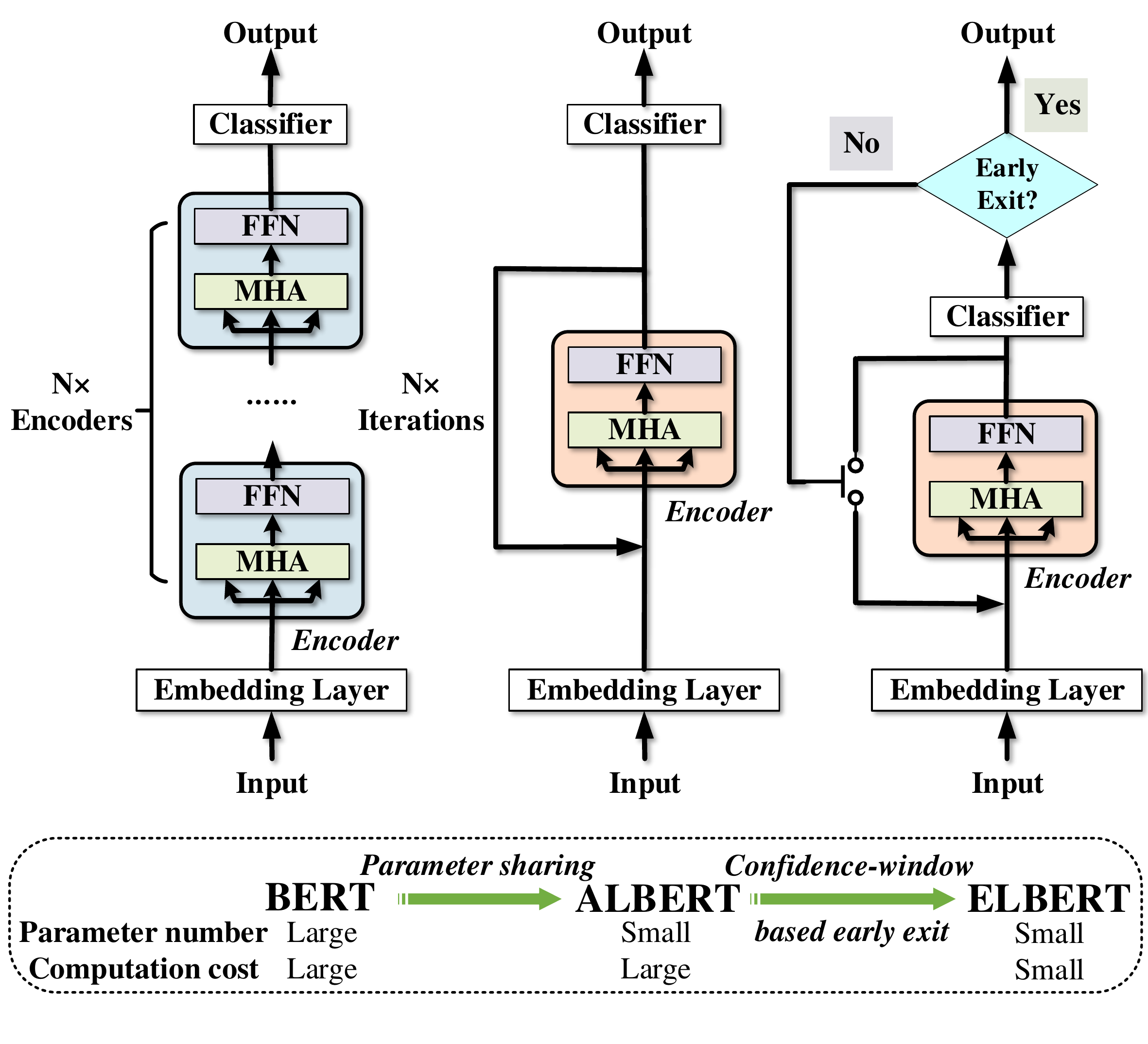}
    \caption{Structures of BERT, ALBERT and ELBERT. Note that ELBERT brings no additional parameters, and the computation brought by early exit mechanism is ignorable\cite{liu2020fastbert} (less than 2\% of that of one encoder). 
    }
    \vspace{-0.1cm}
    \label{fig:Transmodel}
  \end{figure}

\section{Methodology}
\label{sec:method}

  \subsection{Model Arichitecture}
  \label{ssec:model}
  As shown in \myfig{fig:Transmodel}, ELBERT uses ALBERT as the backbone model, which is composed of an encoder and a classifier. 
  Additionally, 
  ELBERT is designed to put a early exit decision after each propagation processed by the encoder and the classifier. 

  \subsection{Training}
  \label{ssec:train}
  To fit the early exit mechanism in inference, the losses of inputs exiting at different depths of ELBERT are calculated during the training. 
  For classification, the early exit loss at the $i$-th layer $\mathcal{L}_i$ is calculated with \emph{Cross-Entropy}
  \begin{equation}
  \mathcal{L}_{i}=-\sum_{c \in C}\left[ \mathbb{I} \left[\mathbf{\hat y}_{i}=c\right] \cdot \log P\left(\mathbf{\hat y}_{i}=c \mid \mathbf{h}_{i}\right)\right], 
  \end{equation}
  where $c$ and $C$ denote one class label and the set of class labels, respectively. 
  The common practice is to simply add up $\mathcal{L}_i$ as the total loss $\mathcal{L}$ \cite{xin2020deebert,liu2020fastbert}. For better training under various combinations of losses, we assign a trainable variable $t_i$ with an initial value of 4 to each layer, inspired by Wang \emph{et al.}\cite{wang2019dynexit}. The weight of the $i$-th layer $w_i$ is calculated by
  \begin{equation} \label{weight}
  w_i=\left\{\begin{array}{ll}\sigma( t_{i}) & 0<i\le M-1 \\ 
  M-\sum_{i=1}^{M-1} \sigma( t_{i}) & i=M\end{array}\right.,
  \end{equation}
  where $M$ denotes the depth of ELBERT and $\sigma(\cdot)$ denotes sigmoid funciton $\sigma\left(t_{i}\right)=1 /\left(1+\exp({-t_{i})}\right)$. Then the total loss $\mathcal{L}$ is calculated by a weighted sum
  \begin{equation}
    \mathcal{L}=\sum_{i=1}^{M}  w_{i}\cdot \mathcal{L}_{i}.
  \end{equation}
  In this way, the cases that inputs may exit at different depths are well considered, which helps to bridge the gap between training and inference of ELBERT. 

  \subsection{Inference}
  \label{ssec:inference}
  ELBERT first introduces a two-stage early exit mechanism, which focuses on both the intermediate state and the historical trend of classifier output to decide whether an early exit of computation is needed. 

  Formally, the input $\mathbf{x}$ goes through the encoder iteratively. The hidden state $\mathbf{h}_i$ after the $i$-th forward propagation of the encoder is calculated by
  \begin{equation}
    \mathbf{h}_i=\left\{\begin{array}{ll}
        Encoder(\mathbf{h}_{i-1}) & 0<i\le M \\ 
        Embedding(\mathbf{x}) & i=0\end{array}\right..
  \end{equation}
  After each forward propagation in the encoder, the hidden state $\mathbf{h}_i$ is sent to the classifier that outputs a prediction probability distribution $p_i=Classifier(\mathbf{h}_i)$ via fully-connected layer and softmax function for classification. Then we can get the predicted label $\mathbf{\hat y}_i = argmax(p_i)$. 

  The first stage of the early exit focuses on confidence, or intermediate state, of the classifier. Given a probability distribution $p_i$, we take its normalized entropy as the $Puzzlement$ of the current classifier, which is calculated by
  \begin{equation}
  Puzzlement(i) = \frac{\sum_{j=1}^{C} p_i(j) \log p_i(j)}{\log (1/C)},
  \end{equation}
  where $C$ denotes the number of labeled classes. The model will stop the inference in advance and take $\mathbf{\hat y}_i$ as the final prediction to skip further computations when $Puzzlement(i) < {\delta}$, where $\delta$ is a user-defined threshold. When a faster model is needed and some accuracy degradation is tolerable, we can set a higher $\delta$, while the opposite situation leads to a lower $\delta$. 

  The second stage traces the historical trend of the classifier output in a time window, whose size $N$ is defined based on user demands. We propose three criteria for triggering the second stage early exit \emph{in a time window}: 1) The prediction probability $p_i$ of a certain class varies monotonically. 2) The range of $max(p_i)$ is less than a set value. 3) The predicted label $\mathbf{\hat y}_i$ stays the same. 
  Experimental results show that the first criterion outperforms others. In subsequent experiments, we will use the first criterion for the second stage by default, and the window size $N$ is set to 8. 

  Usually, we prefer the moment when we get enough confidence. Only when the first stage condition isn't satisfied will we consider the second stage early exit. 

\section{Experiments}
\label{sec:experiments}

  \subsection{Baselines}
  \label{ssec:baseline}
  We select three baselines. 1) Original model: We choose ALBERT-large (depth=24). 2) Plain compression: We evaluate several models with smaller depths based on ALBERT-large. 
  3) Early exit approach: The methods in DeeBERT and FastBERT are applied to ALBERT for comparison. 

  \subsection{Datasets}
  \label{ssec:task}
  To test the generalization ability of ELBERT, widely used GLUE benchmark \cite{wang2019glue}, AG-news \cite{zhang2015character} and IMDB \cite{imdb} are evaluated in our experiments. These datasets include various NLP tasks such as Natural Language Inference, Sentiment Analysis and News Classification. 

  \subsection{Experimental Setup}
  \label{ssec:setup}

  \textbf{Training}\quad For GLUE we use the corresponding hyperparameters in ALBERT original paper for a fair comparison, while for other datasets, we use a default learning rate of 3e-5 and a batch size of 32. 
  \\
  \textbf{Inference}\quad In practical scenes, the user requests often arrive one by one. Our batch size of inference is set to 1, following prior work\cite{xin2020deebert,teerapittayanon2016branchynet}. The experiments are done on an NVIDIA 2080Ti GPU.

  \subsection{Main Results}
  \label{ssec:performance}

\begin{figure}[bt]
        \centering
        \includegraphics[width=8.5cm]{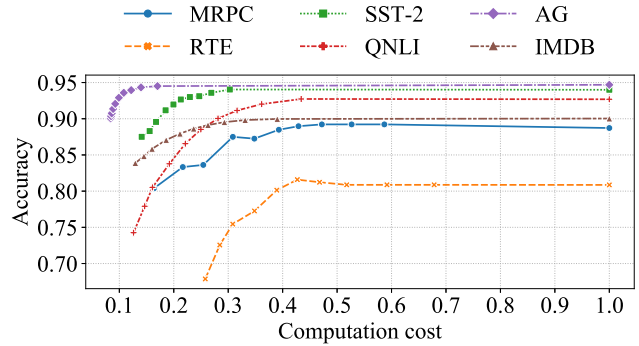}
        \caption{The accuracy-speed tradeoffs of ELBERT on different datasets, where computation cost represents the normalized ratio of original computation. 
        The rightmost point of each curve represents the original model. 
        }
        \label{fig:main_compute_ratio_acc}
  \end{figure}

\begin{figure}[!h]
        \centering
        \begin{minipage}[b]{0.25\linewidth}
          \centerline{\includegraphics[width=8.5cm]{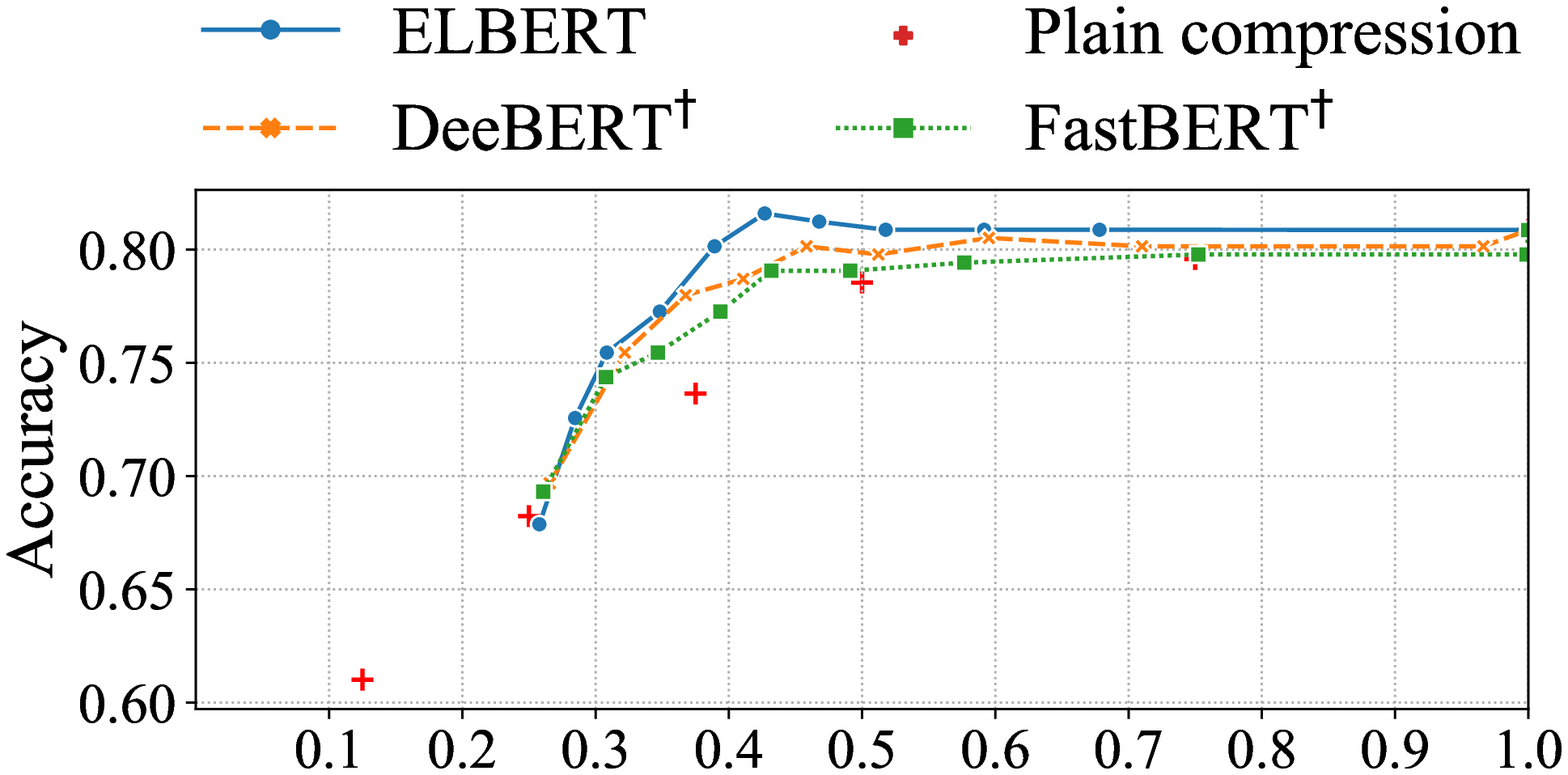}}
        \end{minipage}
        \vfill
        \begin{minipage}[b]{0.25\linewidth}
          \centerline{\includegraphics[width=8.5cm]{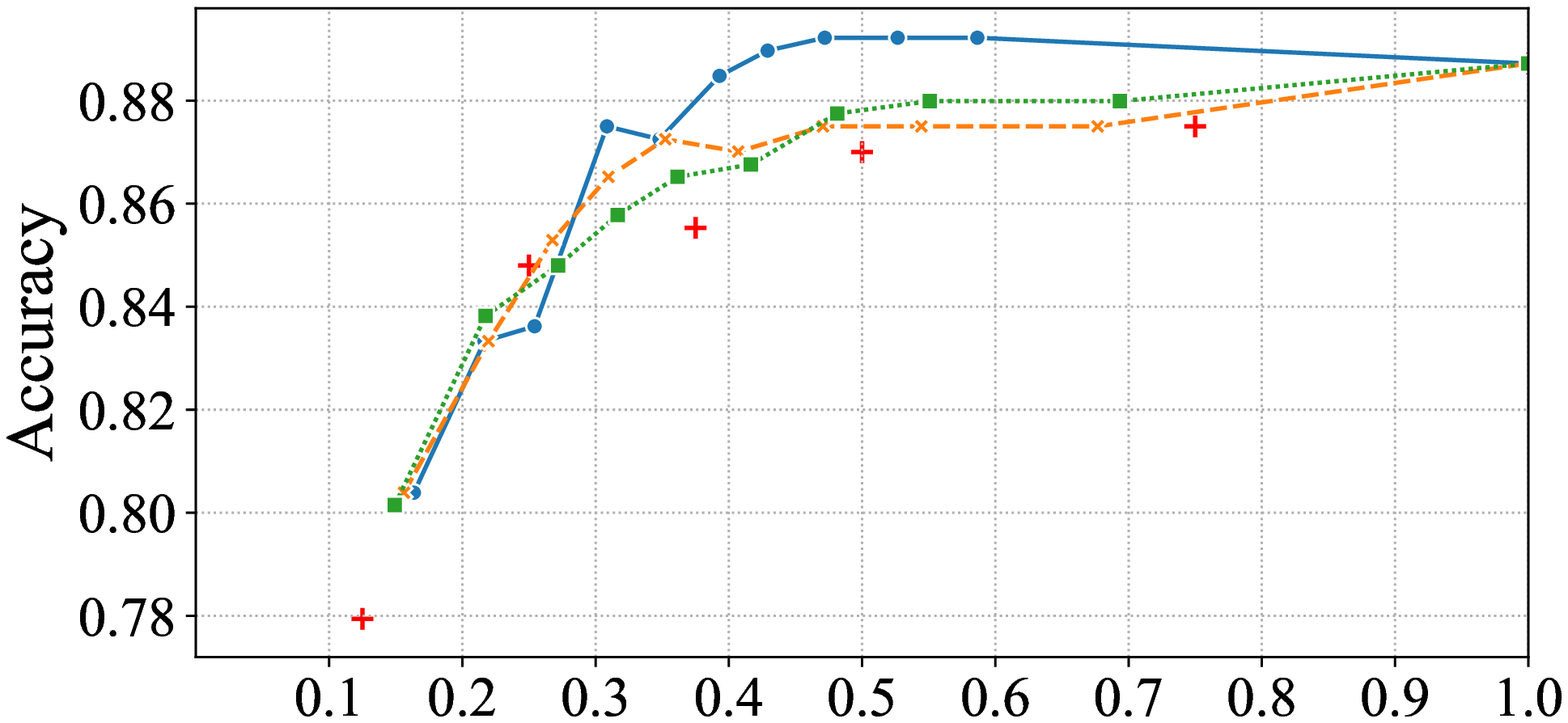}}
        \end{minipage}
        \vfill
        \begin{minipage}[b]{0.25\linewidth}
          \centerline{\includegraphics[width=8.5cm]{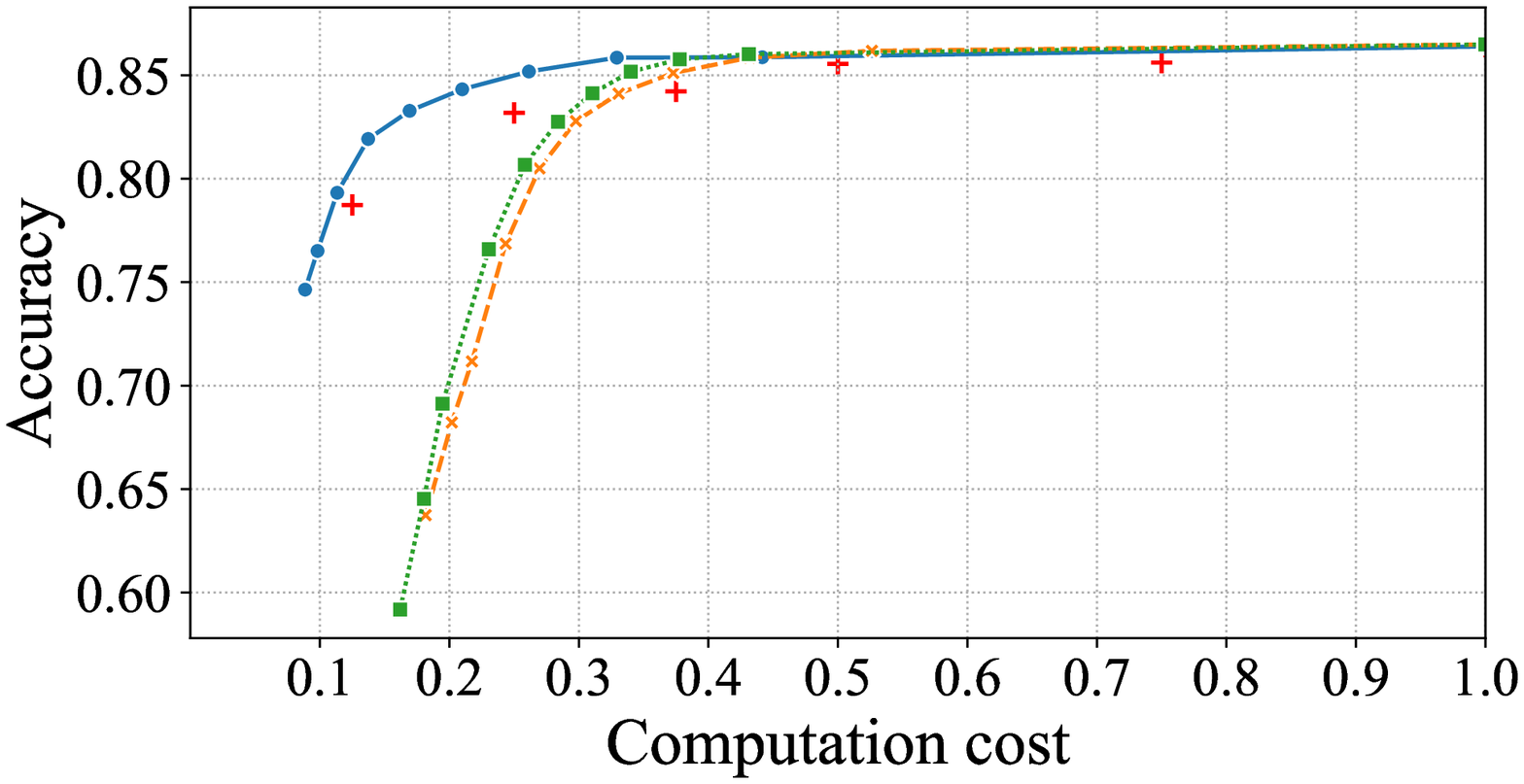}}
        \end{minipage}
        \caption{Comparison on RTE, MRPC and MNLI. 
        }
        \label{fig:comparison}
      \end{figure}

    \textbf{Efficient inference acceleration}\quad We evaluate ELBERT on the above datasets and report the median of 5 runs in \myfig{fig:main_compute_ratio_acc} and \myfig{fig:comparison}. 
    The curves are drawn by interpolating several points that correspond to different $\delta$, which changes from 0.1 to 1.0 with a step size of 0.1 in the first stage of early exit. For all datasets, ELBERT achieves at least two times inference speedup while keeping or even improving the accuracy. When a little accuracy degradation is tolerable, the inference acceleration ratio can be up to ten. This demonstrates ELBERT's superiority of inference acceleration. 
      \\
    \textbf{Task-related trends}\quad An interesting observation is that there are different trends of curves in \myfig{fig:main_compute_ratio_acc} on different kinds of tasks. For News Classification (AG), ELBERT gets the best acceleration performance, followed by Sentiment Analysis (SST-2, IMDB), the curves of which drop a little faster. NLI (QNLI, RTE) is the case with the lowest performance. This indicates that different tasks may have different internal characteristics and acceleration difficulty. Early exit may help us understand tasks better. We will do some discussions about this in \mysec{ssec:see}. 
    \\
    \textbf{Flexible and better accuracy-speed tradeoffs}\quad 
    We compare different models on several datasets. The results are shown in \myfig{fig:comparison}, where the red star-shaped points represent different models obtained by plain compression. 
    Our first observation is that ELBERT significantly outperforms plain compression models. Also, compared to other early exit based methods, ELBERT obtains higher accuracy than both DeeBERT and FastBERT under the same computation cost, which shows ELBERT's great advantages over other approaches.

  \subsection{Visualization of Early Exit}
  \label{ssec:see}
\begin{figure}[t]
          \begin{minipage}[b]{0.49\linewidth}
            \centering
            \centerline{\includegraphics[width=4.0cm]{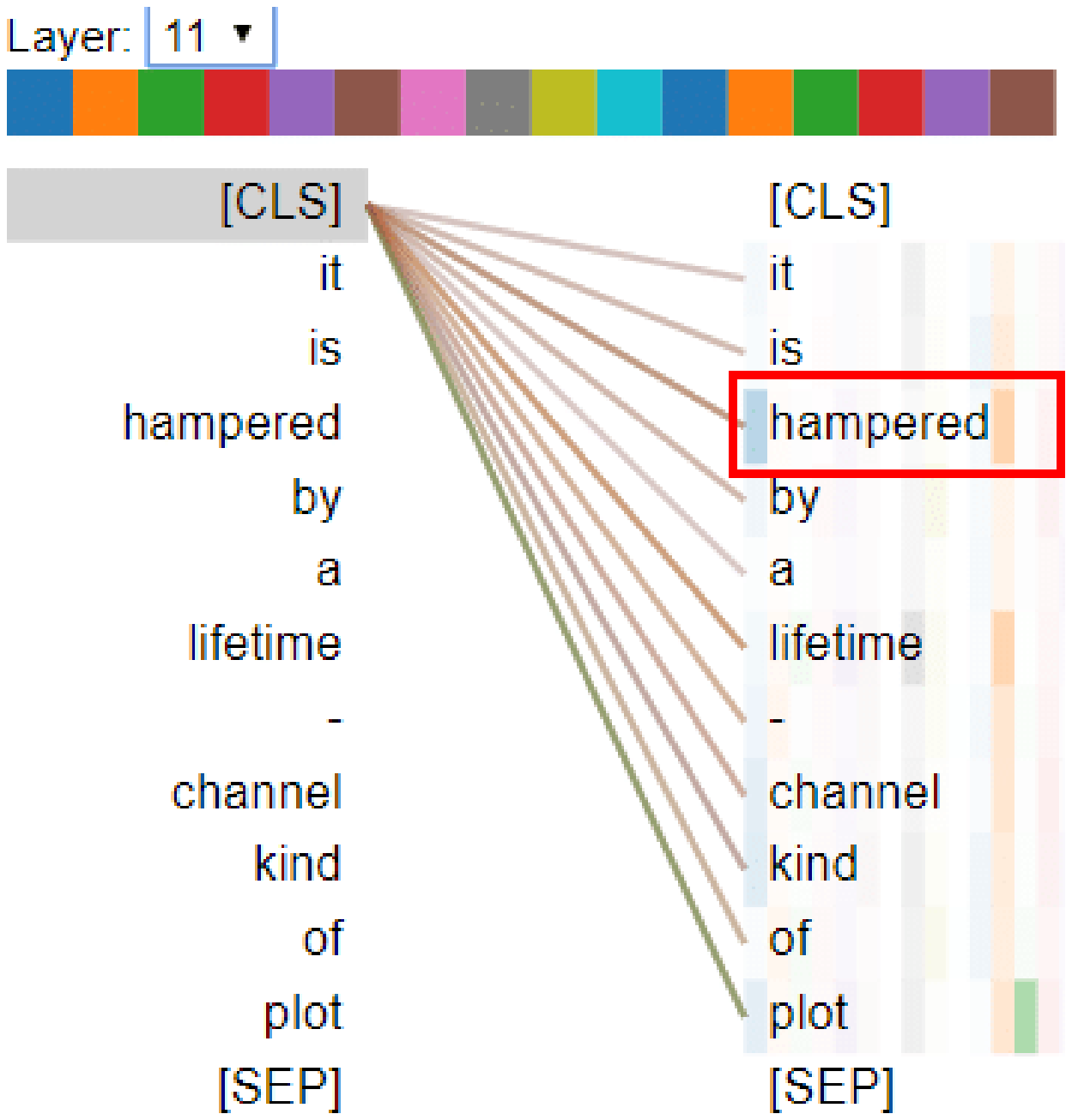}}
            \centerline{(a) Layer 11, $Negative$}\medskip
          \end{minipage}
          \hfill
          \begin{minipage}[b]{0.49\linewidth}
            \centering
            \centerline{\includegraphics[width=4.0cm]{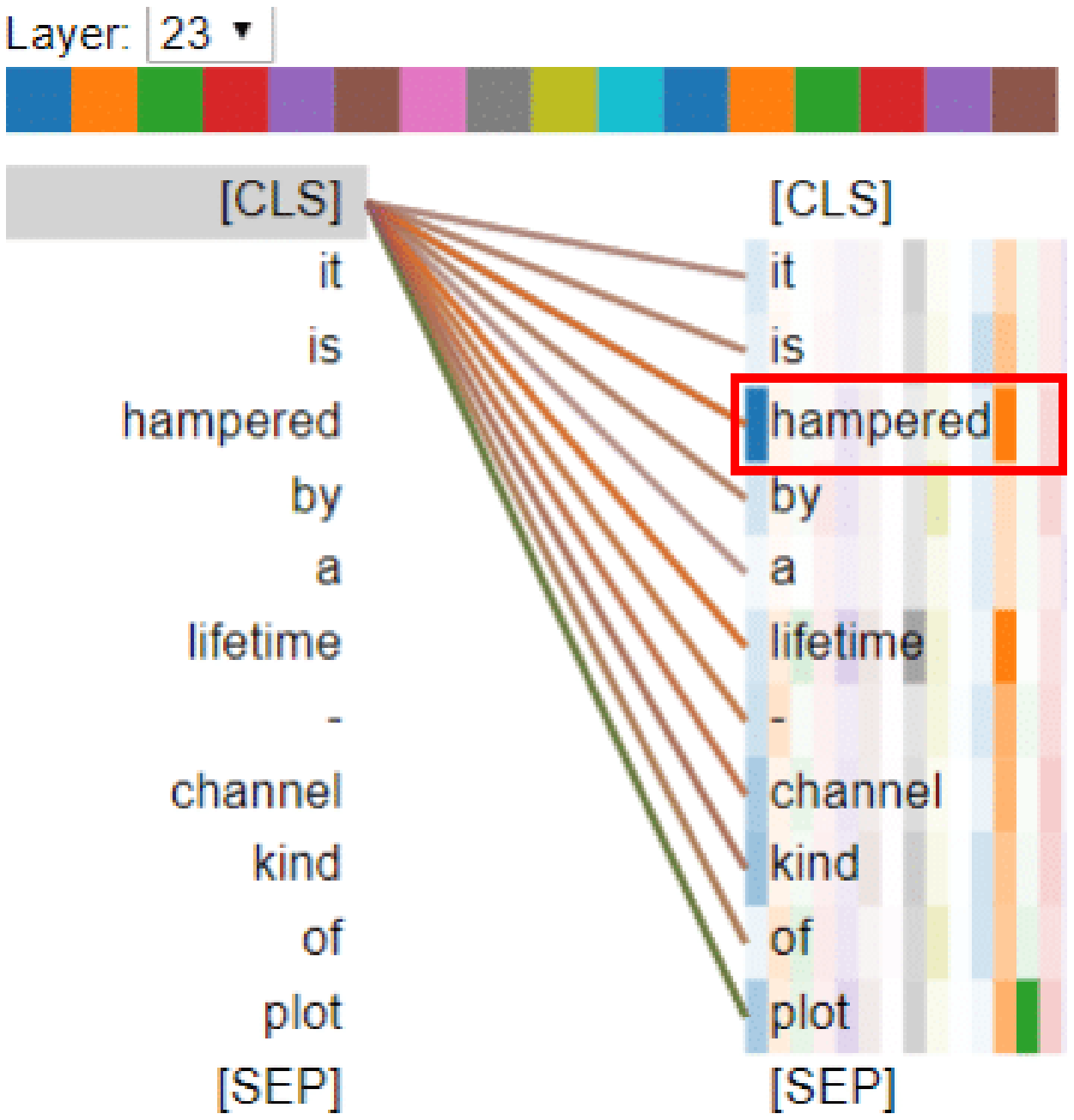}}
            \centerline{(b) Layer 23, $Negative$}\medskip
          \end{minipage}
          \caption{A simple case. The early exit is triggered when the attention to specific word (\textcolor{red}{hampered}) exceeds a certain limit. }
          \label{fig:attn_49}
        \end{figure}

\begin{figure}[t]
      \begin{minipage}[b]{0.49\linewidth}
        \centering
        \centerline{\includegraphics[width=4.0cm]{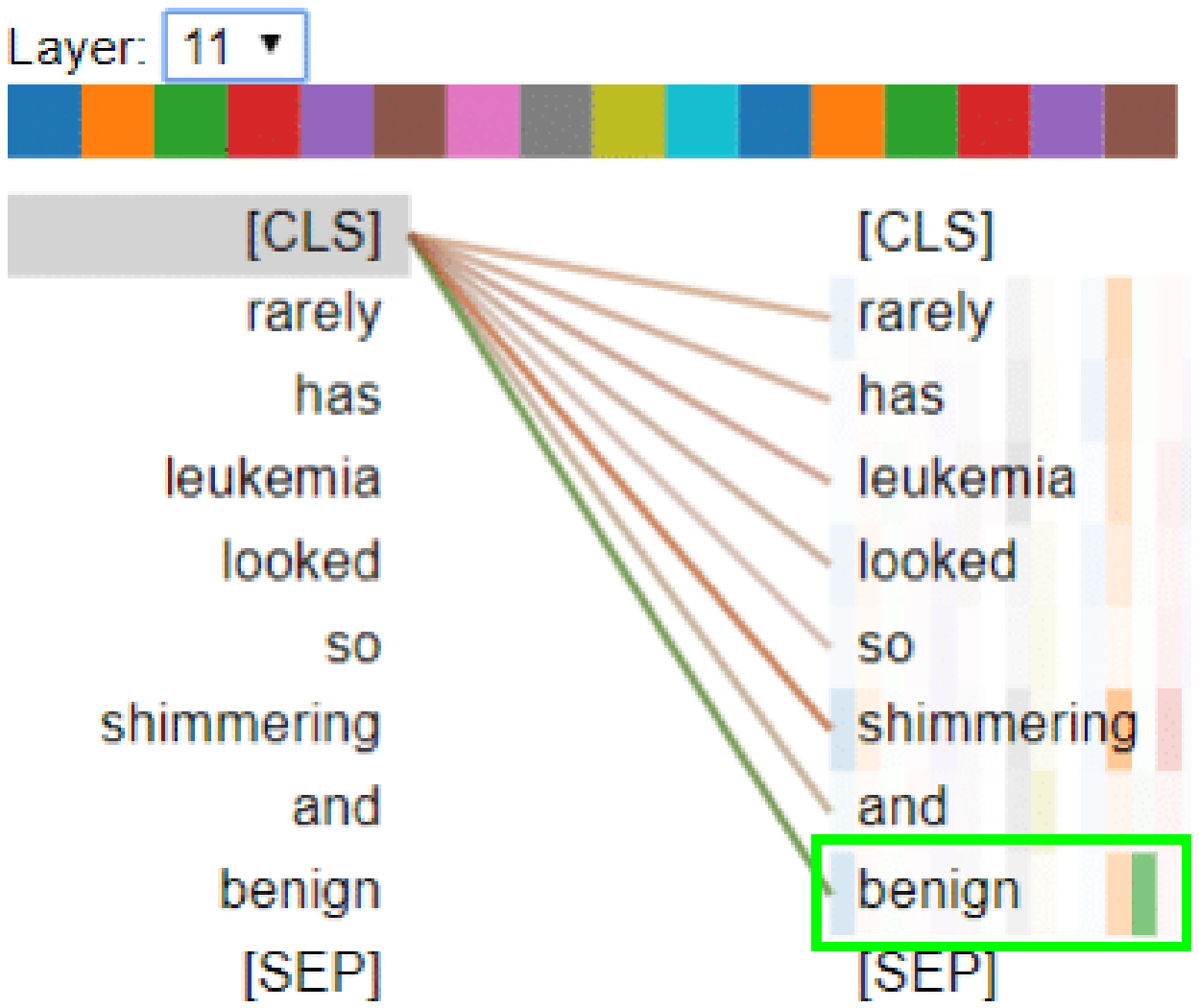}}
        \centerline{(a) Layer 11, $Positive$}\medskip
      \end{minipage}
      \hfill
      \begin{minipage}[b]{0.49\linewidth}
        \centering
        \centerline{\includegraphics[width=4.0cm]{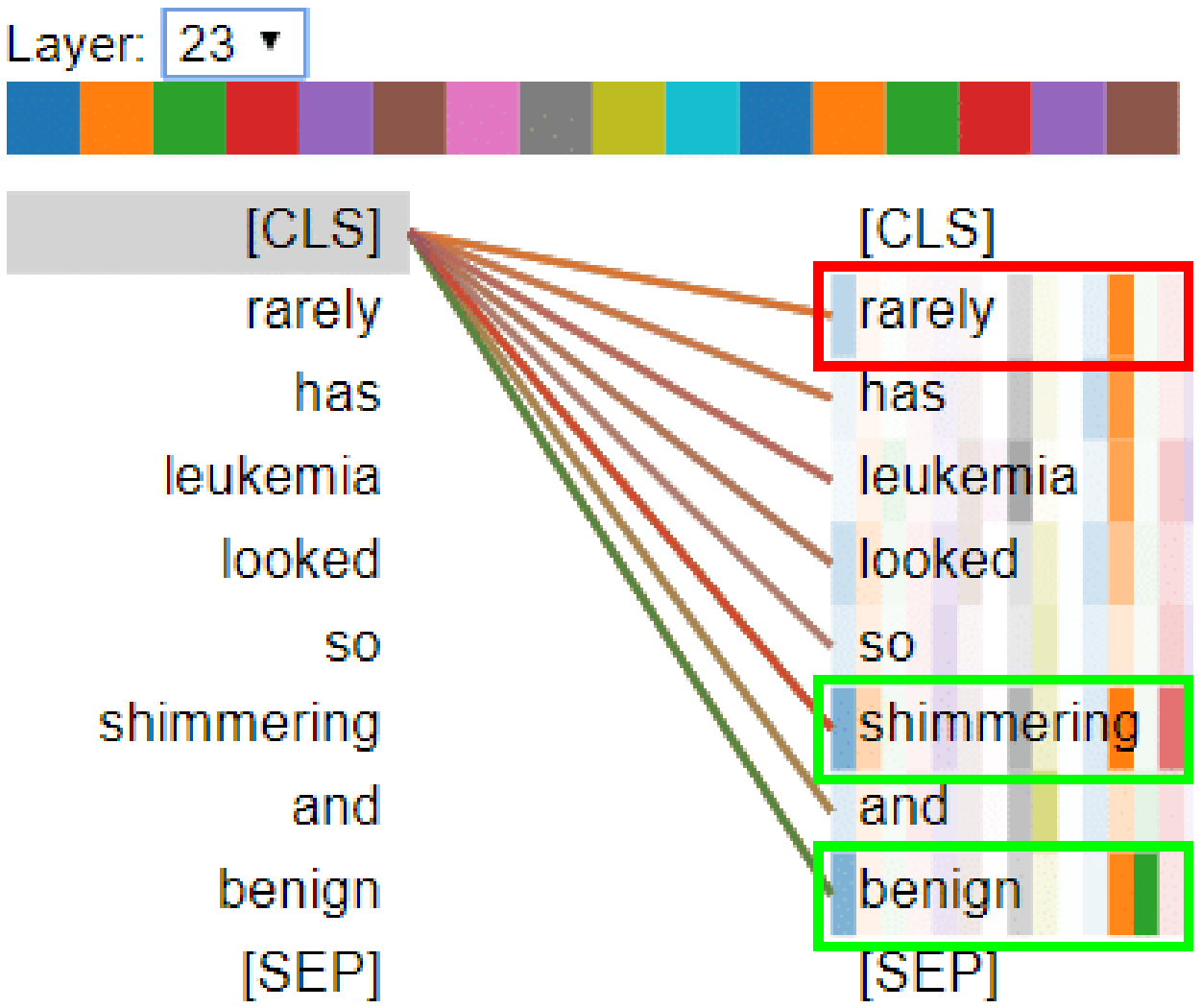}}
        \centerline{(b) Layer 23, $Negative$}\medskip
      \end{minipage}
      \caption{A hard case. The early exit is triggered in time after the commendatory word (\textcolor{green}{benign}) is well noticed, avoiding subsequently overthinking about unrelated negation (\textcolor{red}{rarely}). }
      \label{fig:attn_171}
    \end{figure}

  To visualize the decision-making process of the early exit in ELBERT, we make some changes to BertViz\cite{vig2019bertviz}, a tool for visualizing attention in Transformer. We use the attention-scores of each layer to get the cumulative attention-scores, which allows us to see the attention relationships between tokens clearly as the input passes through different depths of ELBERT. Since ELBERT only takes the [cls] token as the representation of one sentence to do classification, we only show the cumulative attention-scores of [cls] to other tokens in the figures. 
  We take SST, a Sentiment Analysis dataset for example, and find two main patterns of early exit. 
  \let\thefootnote\relax\footnotetext{$\dagger$The results are based on our implementation on ALBERT-large. 
  }
  \\
  \textbf{Simple input, simple exit}\quad For the most common inputs without emotional turns or negative words, as shown in \myfig{fig:attn_49}, the attention of [cls] to the emotional keywords (i.e., hampered) tends to increase monotonously. Early exit is triggered when such attention exceeds a certain limit determined by the $\delta$, thus reducing redundant computations. Actually the prediction remains unchanged after the exit layer 11. 
  \\
  \textbf{Mitigating overthinking}\quad 
  As \myfig{fig:comparison} shows, ELBERT sometimes achieves even higher accuracy than that of the original model, indicating that the early exit mechanism corrects some wrong predictions of the final layer. As shown in \myfig{fig:attn_171}, the model first pays attention to the commendatory word (i.e., benign) and predicts $Positive$. Next, an irrelevant negative word (i.e., rarely) is noticed, seen as the negation of commendatory words. Then the model predicts $Negative$. This is exactly an example of overthinking. 
  In correct cases, the negation and the corresponding word are often noticed simultaneously. 
  
  The above patterns demonstrate that ELBERT's prediction for classification is mainly determined by some key words, such as negatives and those words with strong emotional orientation. The early exit mechanism helps to establish appropriate attention to these words, which enables the model to exit from simple inputs in advance and avoid overthinking. 

\section{Conclusion}
\label{sec:conclusion}
In this paper, we propose ELBERT, a fast ALBERT coupled with a confidence-window based early exit mechanism. Our empirical experiments demonstrate that ELBERT achieves excellent inference acceleration and outperforms other early exit methods used for accelerating BERT. Moreover, it's quite easy for other models to reach fast and flexible inference by using the proposed method. Our future work will include exploring the confidence-window based early exit mechanism on more kinds of models and combining our method with common compression methods. 



\vfill\pagebreak


\bibliographystyle{IEEEbib}
\bibliography{refv1}

\end{document}